\pdfoutput=1

\documentclass[11pt]{article}

\usepackage[final]{acl}
\usepackage{times}
\usepackage{latexsym}

\usepackage[T1]{fontenc}

\usepackage[utf8]{inputenc}

\usepackage{microtype}

\usepackage{inconsolata}

\usepackage{graphicx}

\usepackage{algorithm}
\usepackage{algpseudocode}

\usepackage{amsmath, amssymb}

\usepackage{booktabs}

\usepackage{graphicx}

\usepackage{hyperref}

\usepackage[inline]{enumitem} 
\setlist[itemize]{leftmargin=*, nosep}

\usepackage{graphicx}
\usepackage{subcaption}  
\usepackage{caption}
\usepackage{float}       

\usepackage{longtable}
\usepackage{acronym}
\acrodef{AI}{artificial intelligence}
\acrodef{LLM}{large language model}
\acrodef{SOTA}{state-of-the-art}

\newcommand{\OurModel}[1]{WikiHowAgent}

\definecolor{customgreen}{RGB}{34,139,34} 
\newcommand{\uparrowgreen}{\textcolor{customgreen}{\uparrow}}
\newcommand{\downarrowred}{\textcolor{red}{\downarrow}}

\definecolor{lightpurple}{RGB}{202, 173, 216}
\definecolor{green}{RGB}{176, 217, 146}
\definecolor{lightgreen}{RGB}{236, 247, 177}
\definecolor{pink}{RGB}{233, 182, 190}
\definecolor{blue}{RGB}{140, 203, 234}
\definecolor{lightblue}{RGB}{230, 240, 254}
\definecolor{orange}{RGB}{236, 166, 128}
\definecolor{yellow}{RGB}{243, 225, 156}

\begin{document}
\title{Conversational Reciprocal Learning at Scale: A Multi-LLM Agent Workflow for Procedural Learning and Pedagogic Quality Assessment}

\title{Conversational Education at Scale: A Multi-LLM Agent Workflow for Procedural Learning and Pedagogic Quality Assessment}
\author{Jiahuan Pei\footnotemark[2], Fanghua Ye\footnotemark[3], Xin Sun\footnotemark[4]\footnotemark[5],
  Wentao Deng\footnotemark[6], Koen Hindriks\footnotemark[2], Junxiao Wang\thanks{Corresponding author}\footnotemark[8]  \\
\footnotemark[2] Vrije University of Amsterdam, Amsterdam, The Netherlands \\
\footnotemark[3] University College London, London, United Kingdom \\
\footnotemark[4] University of Amsterdam, Amsterdam, The Netherlands \\
\footnotemark[5] National Institute of Informatics, Tokyo, Japan \\
\footnotemark[6] Shandong University, Qingdao, China \\
\footnotemark[8] Guangzhou University, Guangzhou, China \\ 
\texttt{j.pei2@vu.nl}, \texttt{junxiao.wang@gzhu.edu.cn}\\
}

\maketitle

\begin{abstract}
Large language models (LLMs) have advanced virtual educators and learners, bridging NLP with AI4Education. 
Existing work often lacks scalability and fails to leverage diverse, large-scale course content, with limited frameworks for assessing pedagogic quality. 
To this end, we propose \textbf{\OurModel{}}, a multi-agent workflow leveraging LLMs to simulate interactive teaching-learning conversations. 
It integrates teacher and learner agents, an interaction manager, and an evaluator to facilitate procedural learning and assess pedagogic quality. 
We introduce a dataset of 114,296 teacher-learner conversations grounded in 14,287 tutorials across 17 domains and 727 topics. 
Our evaluation protocol combines computational and rubric-based metrics with human judgment alignment. 
Results demonstrate the workflow's effectiveness in diverse setups, offering insights into LLM capabilities across domains. 
Our datasets and implementations are fully open-sourced.~\footnote{\url{https://github.com/Jiahuan-Pei/WikiHowAgent}}
\end{abstract}


\section{Introduction}
\Acp{LLM} have sparked smart teaching and learning, reshaping education and bridging NLP with AI4Education~\cite{chu2025llm}.
\Acp{LLM} have been employed as single agents in education, serving as teachers~\cite{chen2024empowering,lan2024teachers}, learners~\cite{xu2025classroom}, or evaluators~\cite{hu2025exploring} to enhance personalized learning~\cite{park2024empowering}. 
Specifically, multi-LLMs have shown promise in simulating teacher-learner interactions, assisting competitive problem solving~\cite{islam2024mapcoder}, providing personalized feedback~\cite{park2024empowering}, immersive multimodal training~\cite{pei-2024-arda}, and evaluating educational content~\cite{lagakis2024evaai, jinxin2023cgmi}.
However, none of the existing approaches use diverse and large-scale course content, which is vital for capturing the variability in instructional materials. 
And they lack a comprehensive evaluation for assessing pedagogic quality, limiting insights into the effectiveness of teaching-learning interactions~\cite{xu2025classroom}.

\begin{figure}[!t]
    \centering
    \includegraphics[trim=0cm 0cm 0.3cm 0cm, clip, width=0.9\linewidth]{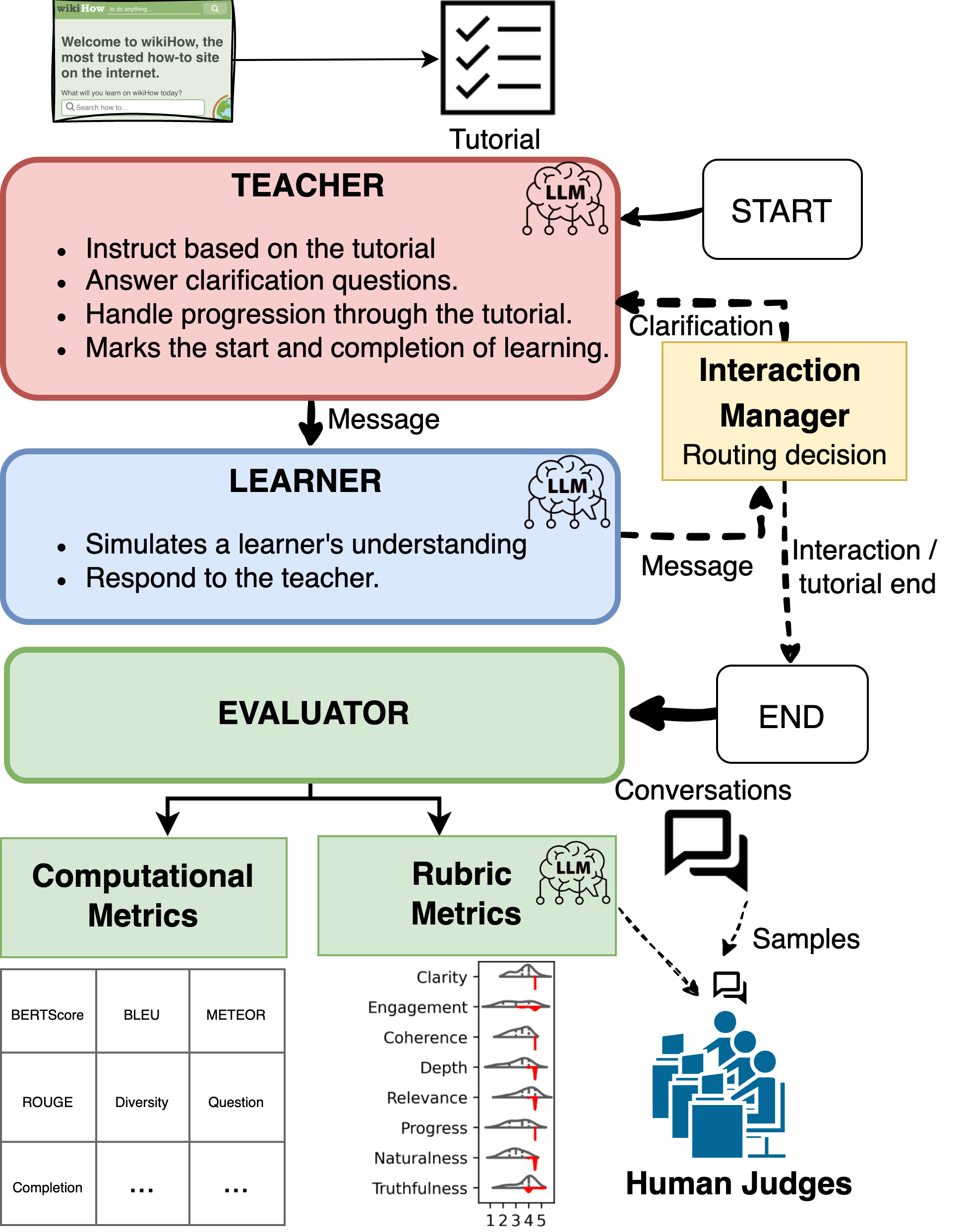}
    \caption{The proposed multi-LLM agent workflow for procedural learning and pedagogic quality assessment.}
    \label{fig:workflow}
\end{figure}

To this end, we introduce a multi-agent workflow, namely \textbf{\OurModel{}} (See Figure~\ref{fig:workflow}), leveraging \acp{LLM} to simulate an interactive teaching-learning conversation graph. 
It utilizes LLM-powered teacher and learner agents, processes instructional tutorials, and generates guided conversations in an interactive mode. 
The workflow is designed to facilitate a seamless interaction between the teacher and learner agents, allowing for a natural flow of conversation.
The workflow consists of four main components: 
\begin{enumerate*}[leftmargin=*, label=(\arabic*)]
    \item \textbf{Teacher Agent}, which provides instructions, answers clarification questions, and guides tutorial progression.
    \item \textbf{Learner Agent}, which simulates understanding and generates responses, either feedback or a question if anything is unclear.
    \item \textbf{Interaction Manager}, which monitors the conversation state, tracks tutorial progress, and determines the next node (i.e., \textsc{teacher}, \textsc{learner}, \textsc{end}) in the graph, ensuring seamless transitions. And 
    \item \textbf{Evaluator}, which assesses the generated conversation using diverse evaluation metrics, including computational metrics and rubric-based metrics generated by an LLM-based evaluator agent, offering insights into pedagogic quality and model performance. 
\end{enumerate*}


We seek the answers to the following research questions:
\begin{enumerate*}[leftmargin=*, label=(\textbf{RQ\arabic*})]
    \item How effective is the proposed multi-agent workflow with diverse \acp{LLM}? How does it perform in homogeneous learning (the same LLM for teacher, learner, and evaluator) versus heterogeneous learning (different LLMs for learner given the same teacher)? 
    \item How do LLMs perform across various domains, and what insights can be drawn about their strengths and limitations in specific areas of procedural learning?
    \item How well do the diverse automatic evaluation metrics align with human judgment in assessing pedagogic quality and conversational performance? 
\end{enumerate*}

We list our key contributions as follows:
\begin{itemize}
    \item A large-scale dataset consisting of 114,296 teacher-learner conversations grounded in 14,287 tutorials. It spans 17 domains and 727 topics, organized in a hierarchical knowledge graph.
    \item A multi-LLM agent workflow for procedural learning and pedagogic quality assessment. A benchmark of 8 prevailing LLMs from diverse providers, including open-source and closed-source models.
    \item A comprehensive evaluation protocol, including diverse metrics and a human judgment alignment assessment method.
    
\end{itemize}

\section{Multi-LLM Agent Procedural Learning}
\begin{algorithm}[t]
\scriptsize
\centering
\caption{Multi-LLM Agent Workflow for Procedural Learning}
\begin{algorithmic}[1]
\Require Tutorial $\mathcal{T} = [t_1, t_2, \dots, t_n]$, max interactions $M$

\State \textbf{Initialize State:}
\State $\mathcal{S} \gets \{\text{messages: } [], \text{tutorial: } \mathcal{T}, \text{current\_step: } 0, $ 
\Statex \hspace{0.72cm} $\text{needs\_clarification: } \text{False}, \text{finished: } \text{False}\}$

\State \textbf{Define Nodes:}
\State \hspace{0.5cm} $\text{TeacherNode: } \mathcal{S} \gets \text{teacher\_agent}(\mathcal{S})$
\State \hspace{0.5cm} $\text{LearnerNode: } \mathcal{S} \gets \text{learner\_agent}(\mathcal{S})$

\State \textbf{Simulate a conversation by iterating through the nodes in the graph:}
\While{$\neg \mathcal{S}.\text{finished}$ \textbf{and} $|\mathcal{S}.\text{messages}| < M$}
    \If{$\mathcal{S}.\text{messages} = []$}
        \State $\mathcal{S} \gets \text{TeacherNode}(\mathcal{S})$ \Comment{Provide first instruction}
    \ElsIf{$\mathcal{S}.\text{messages}[-1] \in $ Teacher}
        \State $\mathcal{S} \gets \text{LearnerNode}(\mathcal{S})$ \Comment{Generate learner response}
        \If{Learner asks a question \textbf{and not} mention ``next step''}
            \State $\mathcal{S}.\text{needs\_clarification} \gets \text{True}$ \Comment{Set flag for Teacher}
        \Else
            \State $\mathcal{S}.\text{current\_step} \gets \mathcal{S}.\text{current\_step} + 1$ \Comment{Next learning step}
        \EndIf
    \Else
        \State $\mathcal{S} \gets \text{TeacherNode}(\mathcal{S})$ \Comment{Provide teacher instruction}
        \If{$\mathcal{S}.\text{needs\_clarification}$}
            \State $\mathcal{S}.\text{needs\_clarification} \gets \text{False}$ \Comment{Reset flag after clarification}
        \EndIf
    \EndIf

    \If{$\mathcal{S}.\text{current\_step} \geq |\mathcal{T}|$}
        \State $\mathcal{S}.\text{finished} \gets \text{True}$ \Comment{End conversation if tutorial is complete}
    \EndIf
\EndWhile

\State \textbf{Evaluate conversation:}
\State $\mathcal{E} \gets \text{Evaluator}(\mathcal{S}.\text{messages})$

\State \Return $\mathcal{S}.\text{messages}, \mathcal{E}$
\end{algorithmic}
\label{alg:multi-llm-agent}
\end{algorithm} 

As shown in Figure~\ref{fig:workflow}, we propose a multi-LLM agent workflow designed for procedural learning through interactive dialogue.
Algorithm~\ref{alg:multi-llm-agent} outlines the implementation of the workflow, comprising a teacher agent, a learner agent, and an evaluator agent, to dynamically simulate guided tutorial sessions using a conversational graph and assess pedagogic quality. 
First, we initialize the learning state and conversation graph, where each node represents a conversational turn, and edges represent interactions between the teacher and learner agents.
Then, we define the nodes, invoking either the teacher agent or the learner agent:
The \textit{teacher agent} is responsible for providing instructions and answering clarification questions. 
The \textit{learner agent} simulates a learner's understanding and generates follow-up questions or responses.
Next, we simulate the conversation by iterating through the nodes in the graph:
\begin{enumerate*}[leftmargin=*, label=(\arabic*)]
    \item It begins with the teacher agent providing instructions based on the first-step's tutorial content.
    \item The interaction manager monitors the conversation state, tracks tutorial progress, handles clarification needs, and determines the next node in the graph.
    \item It continues iteratively, with the teacher agent providing instructions and the learner agent generating responses until the tutorial is completed or a maximum interaction limit is reached.
\end{enumerate*}
Last, the \textit{evaluator} assesses the generated conversation using a range of evaluation metrics, including computational metrics and rubric-based metrics facilitated by an LLM-based evaluator agent.


\section{Evaluation Protocol}\label{sec:evaluation_protocol}
\subsection{Evaluation Metrics}\label{sec:evaluation_metrics}
\subsubsection{Computational Metrics}\label{sec:evaluation_metric_computational}
We utilize a set of automatic metrics, including:
\begin{enumerate*}[leftmargin=*, label=(\arabic*)]
\item \textbf{Question}: Computes the proportion of learner utterances that contain questions.
\item \textbf{Completion}: Computes the proportion of generated conversation reached completion by generating the special token \textsc{finished} in the end.
\item \textbf{Diversity}: Computes a 2-gram diversity score for linguistic variation in responses. 
\item \textbf{BLEU} \citep{papineni-etal-2002-bleu}: Measures the overlap of 4-grams between the generated and reference tutorial, focusing on precision.
\item \textbf{METEOR} \citep{banerjee2005meteor}: Evaluates 4-gram similarity using a harmonic mean of precision and recall, along with alignment of words, including synonyms and stemming.
\item \textbf{ROUGE} \citep{lin2004rouge}: Assesses the quality of generated text by comparing overlapping 4-grams, word sequences, and word pairs with reference tutorial, focusing on recall.
\item \textbf{BERTScore}~\cite{reimers2019}: Measures context-aware semantic similarity between the generated and reference text with BERT embeddings.
\end{enumerate*}

\subsubsection{Rubric Metrics}\label{sec:evaluation_metric_rubric}
We define educational conversation metrics to evaluate tutorial-guided procedural learning and pedagogic quality for both \ac{LLM} and human judges:
\begin{enumerate*}[leftmargin=*, label=(\arabic*)]
\item \textbf{Clarity}: Measures how clear, well-structured, and understandable the teacher's instructions are.
\item \textbf{Truthfulness}: Assesses the factual accuracy and alignment of responses with the tutorial content.
\item \textbf{Engagement}: Evaluates the learner's active participation through meaningful questions and thoughtful responses.
\item \textbf{Coherence}: Checks the logical flow and smooth transitions in the conversation.
\item \textbf{Depth}: Gauges the level of detail and exploration in the discussion beyond surface-level content.
\item \textbf{Relevance}: Ensures responses stay on-topic and align with the tutorial's instructions.
\item \textbf{Progress}: Tracks how effectively the conversation moves forward through the tutorial steps. And
\item \textbf{Naturalness}: Measures how fluid and human-like the conversation feels, avoiding robotic or scripted responses.
\end{enumerate*}
See full rubric in Appendix~\ref{sec:appendix:rubric}.

\subsection{Human Judgment Alignment}
First, we select 25 tutorials based on the top average LLM-judged scores across all metrics, resulting in a set of 175 model-generated conversations.
Second, we collect human annotations of ratings (scale 1-5, where 1 is very poor and 5 is excellent) on the sample set. 
We hire 2 human judges who are master's and PhD students proficient in English and have sufficient learning capabilities.
For each data sample, we have a generated conversation paired with the grounded tutorial. 
We ask each judge to read the tutorial carefully and rate a generated conversation using the tutorial as a reference.
The judges are instructed to evaluate the conversation based on the rubric metrics (\S~\ref{sec:evaluation_metric_rubric}).
Then, we compute the average of the two judges' scores for each metric, using the average value as the reference to evaluate alignment with the automatic metrics.
Last, we conduct a qualitative analysis by visualizing the distribution of ratings provided by human and \ac{LLM} judges, and assess their alignment using the following correlation coefficients:
\begin{enumerate*}[leftmargin=*, label=(\arabic*)]
  \item \textbf{Pearson's $r$}: Evaluates linear correlation between continuous variables.
  \item \textbf{Spearman's $\rho$}: Measures monotonic relationships between ranked variables. And 
  \item \textbf{Kendall's $\tau$}: Assesses ordinal associations between ranked variables.
\end{enumerate*}

\section{Dataset}\label{sec:dataset}
\begin{figure}[htb!]
    \centering
    \includegraphics[trim=1cm 0cm 1cm 0cm, clip, width=\columnwidth]{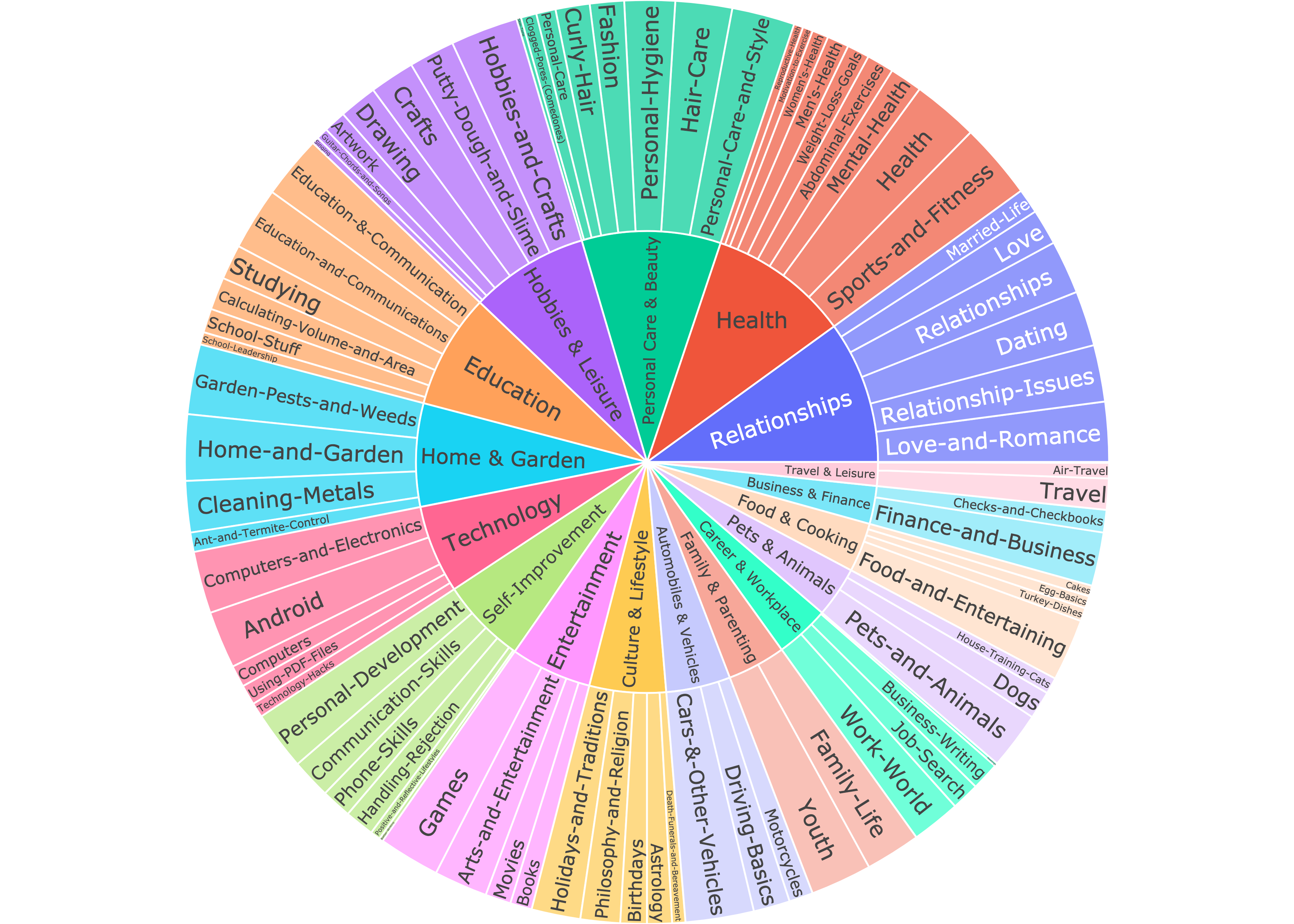}
    \caption{Hierarchical distribution of topics across 17 domains with corresponding topic counts. The inner ring represents the domains, while the outer segments depict 81 first-level topics within each domain. Segment sizes are proportional to the count of entries per topic.}
    \label{fig:tasks_per_topic}
\end{figure}

\begin{table*}[htb!]
\centering
\scriptsize
\resizebox{\textwidth}{!}{
\setlength{\tabcolsep}{3pt} 
\begin{tabular}{@{}llccccccc@{}}
\toprule
\textbf{LLMs} & \textbf{Provider} & \textbf{BERTScore (\%)} & \textbf{BLEU (\%)} & \textbf{METEOR (\%)} & \textbf{ROUGE (\%)} & \textbf{Diversity (\%)} & \textbf{Question (\%)} & \textbf{Completion (\%)} \\ \midrule
DeepSeek & DeekSeek          & 71.81                   & 31.54              & 30.53                & 37.71               & 77.24                         & 17.67                        & 99.57                          \\
Qwen2    & Alibaba           & 73.29                   & 37.48              & 32.06                & 41.96               & 83.44                         & 20.51                        & 99.49                          \\
Gemma    & Google            & 66.85                   & 28.92              & 22.16                & 36.13               & 85.39                         & 27.99                        & 97.17                          \\
OLMo2    & Allenai           & 71.45                   & 25.42              & 32.22                & 34.25               & 79.80                         & 22.75                        & 100.00                         \\
OpenChat & Tsinghua          & 71.06                   & 29.33              & 33.66                & 37.62               & 73.31                         & 62.36                        & 79.46                          \\
Llama3   & Meta              & 74.74                   & 35.27              & 39.33                & 44.85               & 75.66                         & 61.60                        & 86.39                          \\
Phi4     & Microsoft         & 64.46                   & 28.42              & 34.90                & 37.44               & 79.33                         & 57.27                        & 99.43                          \\
GPT-4    & Open AI          &  69.66                        &  43.86                  & 34.48                     & 48.58                    & 86.96                              & 60.82                             &  99.79                              \\ \midrule
\textbf{Overall}  & $\mu \pm \delta $                & 70.42\tiny{$\pm$3.36}  & 32.53\tiny{$\pm$6.00} & 32.42\tiny{$\pm$4.91} & 39.82\tiny{$\pm$3.62}    & 80.14\tiny{$\pm$4.79}        &  38.59\tiny{$\pm$20.70}        & 95.16\tiny{$\pm$7.82}          \\ 
\bottomrule
\end{tabular}
}
\caption{Overall workflow effectiveness in homogeneous learning (same LLM for teacher, learner, and evaluator), benchmarking 8 prevailing \acp{LLM} as backbone, evaluated on 7 computational evaluation metrics. Symbols $\mu$ and $\delta$ denote the mean and standard deviation of the evaluation metrics, respectively.}
\label{tab:benchmark_homo}
\end{table*}
We collected 3,680 real-world instructional articles from WikiHow\footnote{\url{https://www.wikihow.com/}} covering 17 domains and 727 topics, as shown in Figure~\ref{fig:tasks_per_topic}.
Each article has 1516.7 tokens and 4.0 methods as how-to tutorials on average.
Each method contains 304.3 tokens and 5.2 instructional steps on average. 
The data collection contains 14,738 methods and 49,344 learning steps in total.
We treat each method as a tutorial and generate 14,287 teacher-learner conversations, skipping tutorials with only one step, for 8 backbone \acp{LLM}, resulting in a total of 114,296 conversations.
Additionally, we count the number of articles for each domain (Appendix~\ref{sec:article_per_domain}) and organize topics into a hierarchical knowledge graph (Appendix~\ref{sec:appendix:kg}).

\section{Results}\label{sec:results}
We conduct experiments with 8 \acp{LLM} from diverse providers: The open-source models including DeepSeek (7B), Qwen2 (7B), Gemma (7B), OLMo2 (7B), OpenChat (7B), Llama3 (8B), Phi4 (14B), and commerical GPT-4 (1.76TB).

\subsection{Evaluation Results of the Workflow (RQ1)}\label{sec:performance_across_models}

\subsubsection{Homogeneous Learning}
In this setting, all three agents (teacher, learner, and evaluator) share the same model.
As shown in Table~\ref{tab:benchmark_homo}, we present the evaluation results of the proposed multi-LLM agent workflow with 8 backbone \acp{LLM} in homogeneous learning, evaluated by computational metrics.
Overall, the workflow shows strong effectiveness in homogeneous learning setups, with high diversity and completion rates.
Specifically, we find that:

First, the proposed workflow achieves consistently outperforming mean scores in \textbf{Completion} ($\mu=95.16\%$), \textbf{Diversity} ($\mu=80.14\%$), and semantic similarity (\textbf{BERTScore} $\mu=70.42\%$).
This indicates the workflow's ability to generate varied and complete conversations that are semantically aligned with the reference tutorial. 
This is attributed to the instruction-following capabilities of \acp{LLM}, which enable the workflow to complete conversations with diverse yet coherent responses, while the teacher agent effectively mirrors the tutorial's instructional content. 

Second, simulating learner engagement for educational conversation is still challenging.
OpenChat and Llama3 excel in \textbf{Question} generation ($\mu=$ 62.36\%, 61.60\%), showcasing their capability to simulate active learner engagement, 
but the workflow demonstrates variability in the learner agent's performance across scenarios, as indicated by the high standard deviation in the \textbf{Question} ($\delta=20.70\%$).
This variability may stem from the diverse nature of the tutorials, which can influence the learner agent's engagement and question generation.
Besides, the task to understand the teacher's instructions and generate questions is challenging, leading to variant performance on different \acp{LLM}. 

Third, the workflow demonstrates variability in alignment with the reference tutorial, as reflected by the high standard deviations in \textbf{BLEU}, \textbf{METEOR}, \textbf{ROUGE}, and \textbf{BERTScore} ($\delta=6.00, 4.91, 3.62, 3.36$) metrics. 
This suggests that the performance is sensitive to the specific content and structure of the tutorial, as these metrics heavily rely on the overlap and semantic similarity between the generated conversation and the reference tutorial. 
The observed deviations highlight the challenges in maintaining consistent quality across diverse tutorial topics and instructional styles.

\begin{table*}[htb!]
\centering
\resizebox{\textwidth}{!}{
\setlength{\tabcolsep}{3.0pt} 
\begin{tabular}{@{}lccccccc@{}}
\toprule
\textbf{Learner} & \textbf{BERTScore (\%)} & \textbf{BLEU (\%)} & \textbf{METEOR (\%)} & \textbf{ROUGE (\%)} & \textbf{Diversity (\%)} & \textbf{Question (\%)} & \textbf{Completion (\%)} \\ 
\midrule
OpenChat             & 71.06                   & 29.33              & 33.66                & 37.62               & 73.31                   & 62.36                  & 79.46  \\ \midrule
DeepSeek             & 68.08 $\downarrowred$                  & 33.73 $\uparrowgreen$              & 30.49 $\downarrowred$               & 39.06 $\uparrowgreen$              & 77.79 $\uparrowgreen$                  & 26.87 $\downarrowred$                 & 99.30 $\uparrowgreen$                   \\
Qwen2                & 71.85 $\uparrowgreen$                  & 37.08 $\uparrowgreen$             & 32.18 $\downarrowred$               & 42.96 $\uparrowgreen$              & 80.90 $\uparrowgreen$                  & 26.21 $\downarrowred$                 & 98.27 $\uparrowgreen$                   \\
Gemma                & 68.99 $\downarrowred$                  & 36.75 $\uparrowgreen$             & 31.19 $\downarrowred$               & 42.39 $\uparrowgreen$              & 84.50 $\uparrowgreen$                  & 23.59 $\downarrowred$                 & 98.99 $\uparrowgreen$                   \\
OLMo2                & 68.22 $\downarrowred$                  & 35.44 $\uparrowgreen$             & 30.61 $\downarrowred$               & 40.86 $\uparrowgreen$              & 80.81 $\uparrowgreen$                  & 42.11 $\downarrowred$                 & 99.32 $\uparrowgreen$                   \\
\toprule
\textbf{Homogeneous Overall} ($\mu \pm \delta$)                & 70.42\tiny{$\pm$3.36}  & 32.53\tiny{$\pm$6.00} & 32.42\tiny{$\pm$4.91} & 39.82\tiny{$\pm$3.62}    & 80.14\tiny{$\pm$4.79}        &  38.59\tiny{$\pm$20.70}        & 95.16\tiny{$\pm$7.82}            \\ 
\textbf{Heterogeneous Overall} ($\mu \pm \delta$)  & 69.29 $\downarrowred$ \tiny{$\pm$\textit{1.76}} & 35.75$\uparrowgreen$\tiny{$\pm$\textit{1.52}} & 31.12 $\downarrowred$ \tiny{$\pm$\textit{0.77}} & 41.32$\uparrowgreen$\tiny{$\pm$\textit{1.75}} & 81.00$\uparrowgreen$\tiny{$\pm$\textit{2.75}} & 29.70$\downarrowred$\tiny{$\pm$\textit{8.40}} & 98.97$\uparrowgreen$\tiny{$\pm$\textit{0.49}}                     \\
\bottomrule                
\end{tabular}
}
\caption{Overall workflow effectiveness in heterogeneous learning (diverse LLMs as learners with the same LLM as the teacher and evaluator).
Symbols $\mu$ and $\delta$ denote the mean and standard deviation of the evaluation metrics, respectively. 
Symbols $\uparrowgreen$ and $\downarrowred$ indicate an increase or decrease compared to the baseline setting, respectively.}\label{tab:benchmark_heter}
\end{table*}
\begin{figure}[htb!]
    \centering
    \includegraphics[trim=0.8cm 3.5cm 0.3cm 3.5cm, clip, width=0.95\columnwidth]{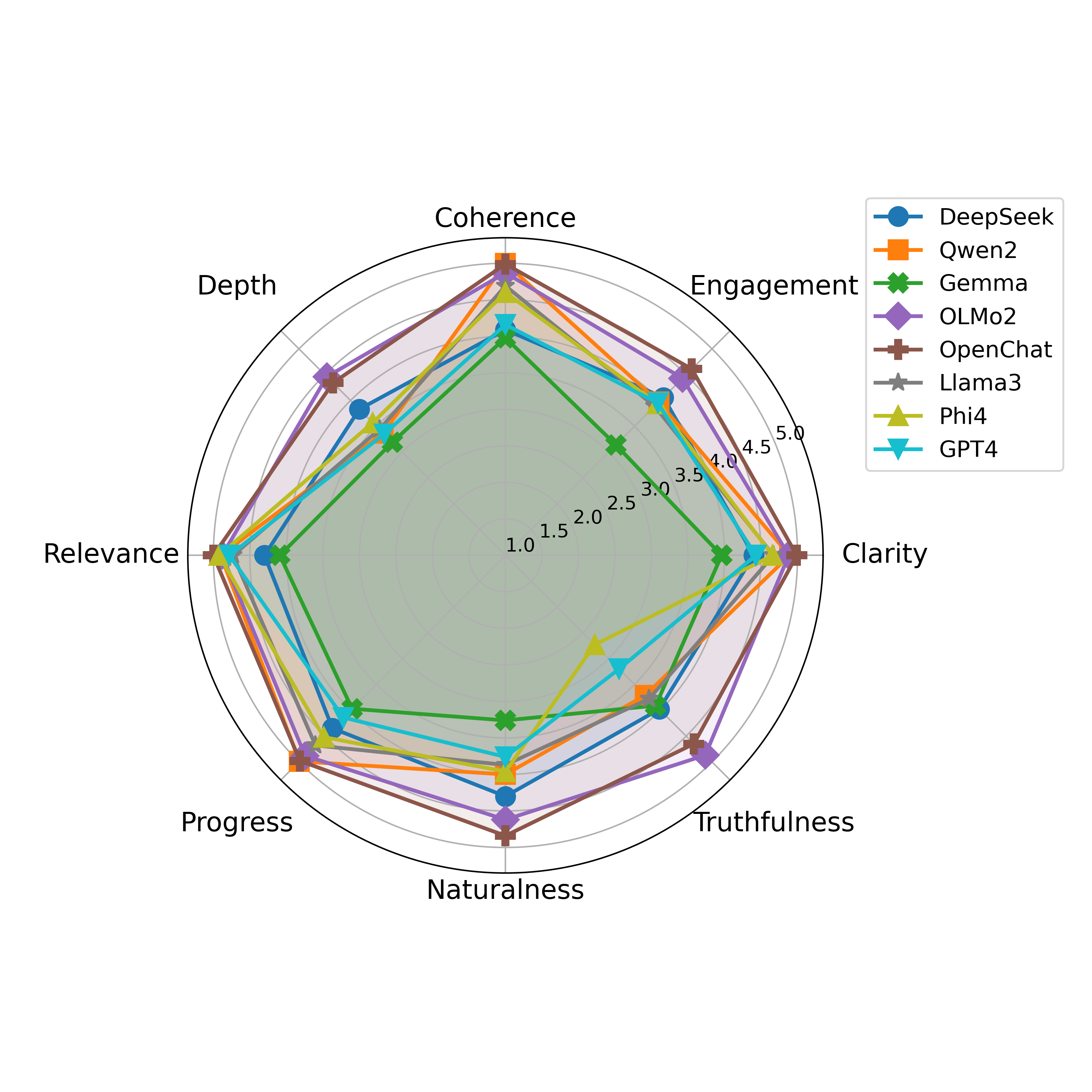}
    \caption{Evaluation results of homogeneous learning using rubric-based metrics across different LLMs.}
    \label{fig:benchmark_homo}
\end{figure}
As shown in Figure~\ref{fig:benchmark_homo}, we visualize the evaluation results of the proposed multi-LLM agent workflow with 8 backbone \acp{LLM} in homogeneous learning, evaluated using rubric-based metrics.
We obtain the following key insights:
\begin{enumerate*}[leftmargin=*, label=(\arabic*)]
\item The workflow has a strong ability to stay on-topic (\textbf{Relevance} $\mu=4.70$) and provide clear instructions (\textbf{Clarity} $\mu=4.61$).
\item Some models struggle to maintain conversational fluidity and explore concepts in depth, aligning with the reference tutorial. Specifically, \textbf{Truthfulness}, \textbf{Naturalness}, and \textbf{Depth} metrics show high variability ($\delta=0.7, 0.5, 0.47$).
These findings highlight the need for further improvements in generating conversations that are human-like, detailed, and factually accurate while closely adhering to the tutorial content.
\item OpenChat and OLMo2 are the most well-rounded performers, excelling in the majority of the metrics. 
Hence, in the heterogeneous learning setups, we equip OpenChat with the teacher and evaluator agents and the other \acp{LLM} for diverse learners.
\end{enumerate*}

\subsubsection{Heterogeneous Learning}
In this setting, the learner agent differs from the teacher and evaluator agents. 
OpenChat serves as the teacher and evaluator, while the learner is selected from 7B open-source \acp{LLM} (DeepSeek, Qwen2, Gemma, OLMo2). 
In Table~\ref{tab:benchmark_heter}, we present the computational evaluation results and observe:

First, the workflow demonstrates improved linguistic \textbf{Diversity} ($+0.86\%$) and alignment metrics, (\textbf{BLEU} $+3.23\%$, \textbf{ROUGE} $+1.50\%$), while maintaining high \textbf{Completion} rates ($\mu=98.97\%$). 
This highlights the benefits of diverse learner agents in generating varied, complete, and well-aligned conversations. 
Additionally, reduced standard deviations across metrics indicate more consistent performance in heterogeneous setups.

Second, learner engagement in heterogeneous setups significantly drops compared to homogeneous setups, as indicated by the \textbf{Question} metric ($-8.89\%$).
This suggests that diverse learners may struggle to generate meaningful questions consistently.
We attribute this to inherent differences in the models' design artifacts (e.g., whether reasoning capabilities are enabled) and data (e.g., domain knowledge and linguistic diversity).
The high standard deviation in the \textbf{Question} metric ($\delta=20.70\%$) further highlights variability in learner engagement across different models.
Addressing this gap requires incorporating adaptive dialogue strategies and human-in-the-loop evaluations to better simulate realistic learning scenarios and improve alignment between model-generated interactions and actual learner needs.

Third, the workflow's performance in semantic similarity (\textbf{BERTScore} $-1.13\%$ and \textbf{METEOR} $-1.30\%$) slightly decreases compared to the homogeneous setting, 
This indicates challenges in maintaining semantic alignment and precision-recall balance with the reference tutorial.

\begin{figure}[htb!]
    \centering
    \includegraphics[trim=0.8cm 3.5cm 0.3cm 3.5cm, clip, width=0.95\columnwidth]{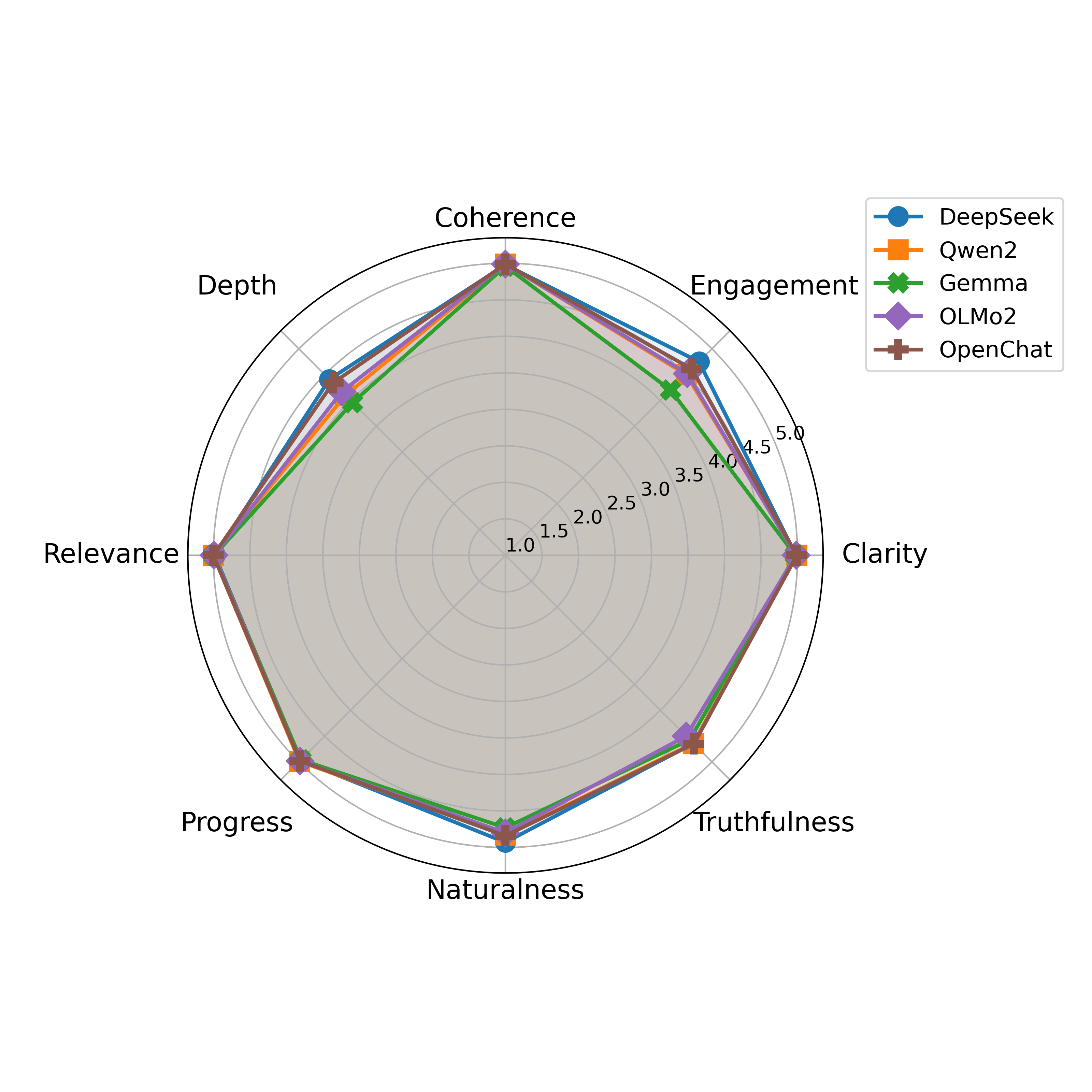}
    \caption{Evaluation results of heterogeneous learning using rubric-based metrics across different LLMs.}
    \label{fig:benchmark_heter}
\end{figure}

\begin{figure*}[t]
    \centering
    \includegraphics[width=0.9\linewidth]{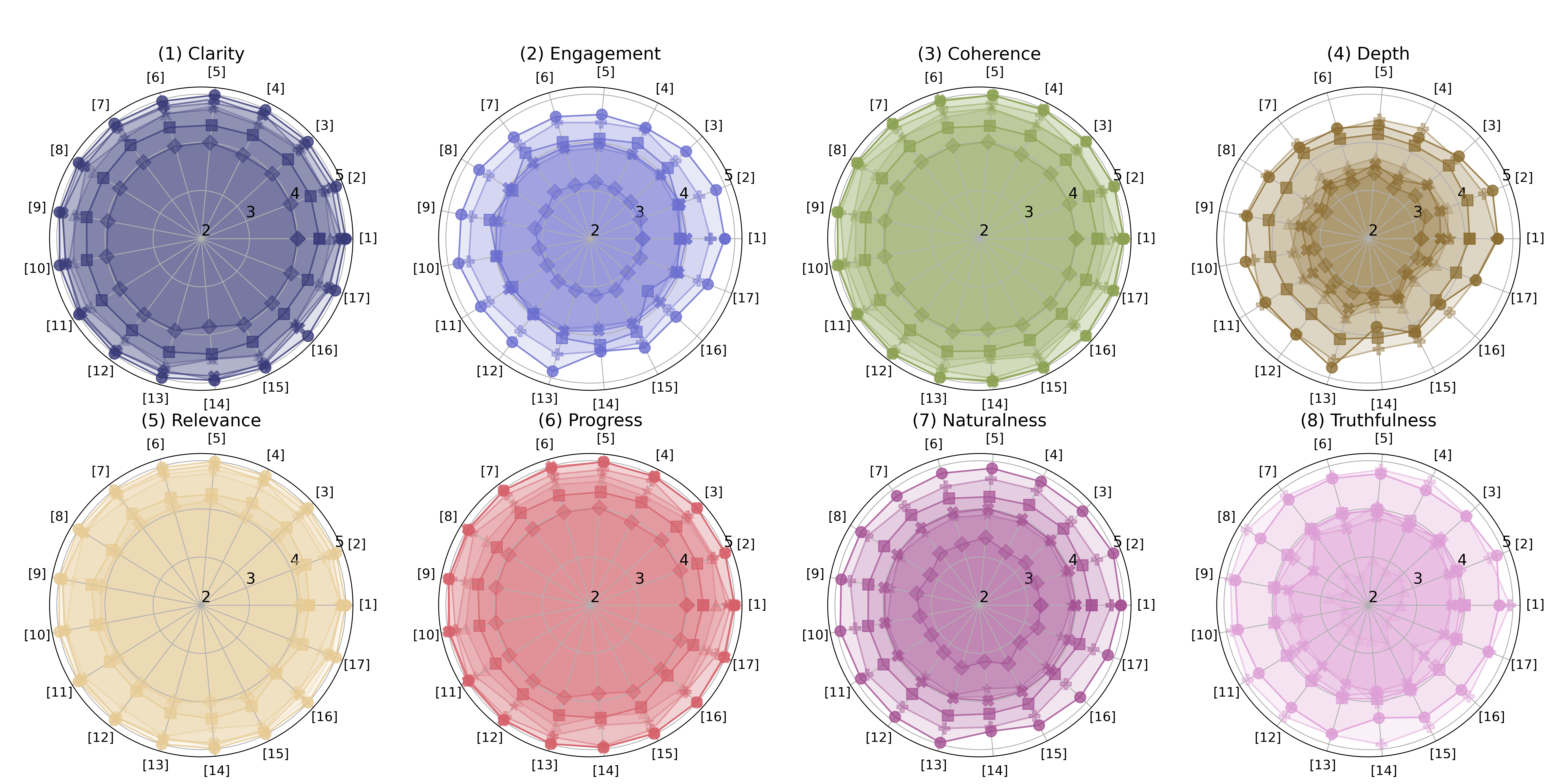}
      \caption{Multiple rubric evaluation scores of all \acp{LLM} over various domains (marked by their IDs in Appendix~\ref{sec:article_per_domain}).}
    \label{fig:benchmark_domain}
\end{figure*}

Last, comparing Figure~\ref{fig:benchmark_heter} with Figure~\ref{fig:benchmark_homo}, we intuitively observe that the performance in heterogeneous learning is more consistent across different models.
However, noticeable variations persist in \textbf{Depth} and \textbf{Engagement} metrics.

\subsection{Performance across Domains (RQ2)}\label{sec:benchmark_domain}

We evaluate the performance of the proposed multi-LLM agent workflow across 17 domains, as shown in Figure~\ref{fig:benchmark_domain}.
We present the evaluation results of the proposed multi-LLM agent workflow with 8 backbone \acp{LLM} in heterogeneous learning, evaluated by computational metrics.
Overall, the workflow shows strong effectiveness in heterogeneous learning setups, with high diversity and completion rates.
Specifically, we observe the following key findings:

On one hand, the proposed workflow achieves consistently well-rounded performance for \textbf{Clarity}, \textbf{Coherence}, \textbf{Relevance} and \textbf{Progress} for all models across all domains.
This indicates the workflow's ability to generate varied and complete conversations that are semantically aligned with the reference tutorial.
This is attributed to the instruction-following capabilities of \acp{LLM}, which enable the workflow to complete conversations with diverse yet coherent responses, while the teacher agent effectively mirrors the tutorial's instructional content.

On the other hand, \textbf{Engagement}, \textbf{Truthfulness}, \textbf{Depth}, and \textbf{Naturalness} exhibit significant variability across domains and different \acp{LLM}. 
This variability likely arises from the diverse complexity and structure of the tutorials, which can affect the learner agent's ability to generate meaningful questions and responses. 
Moreover, the inherent challenge of interpreting the teacher's instructions and formulating appropriate follow-up questions further contributes to the observed performance disparities among different \acp{LLM}.

\subsection{Human Evaluation Alignment (RQ3)}\label{sec:human_alignment}
We measure the alignment between automatic evaluation metrics (\S~\ref{sec:evaluation_protocol}) and human annotations.

\begin{figure}
    \centering
    \includegraphics[width=\columnwidth]{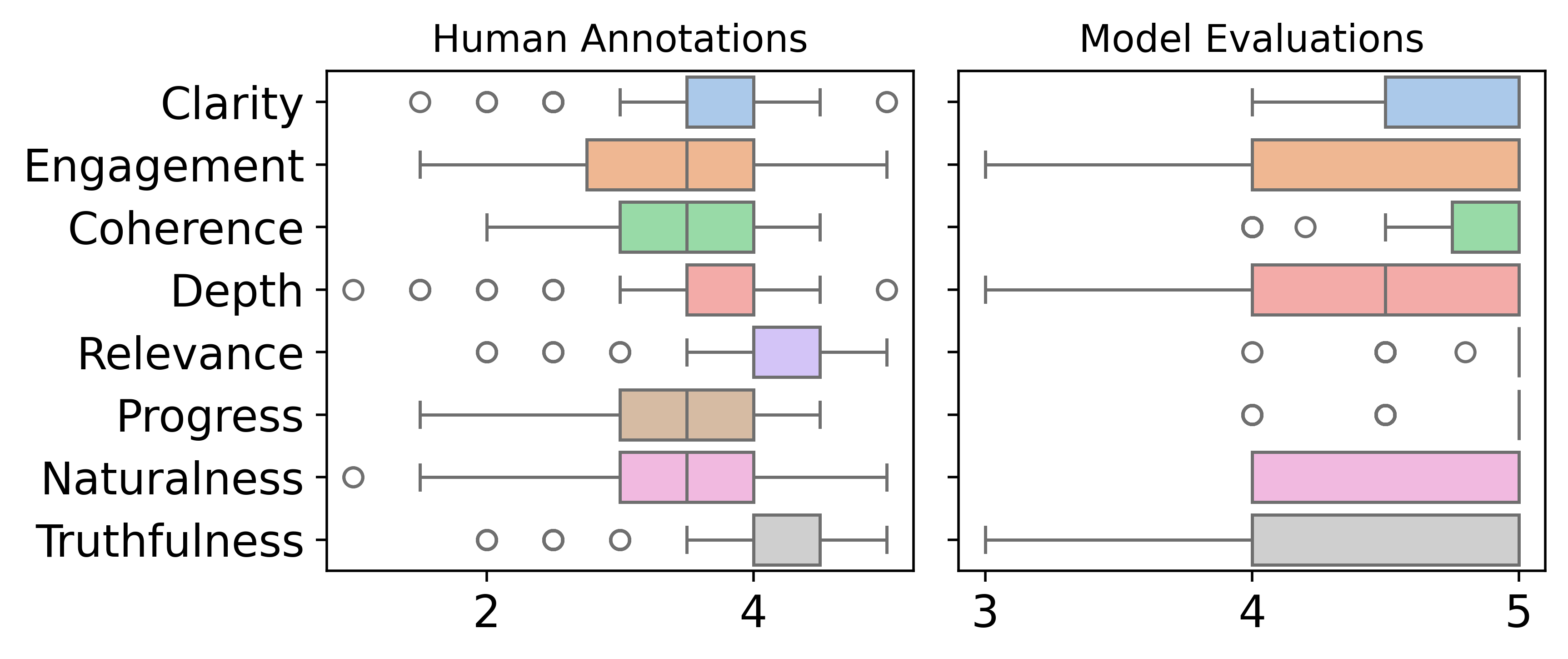}
    \caption{Comparison between human annotation and model evaluation scores.}
    \label{fig:human_alignment_boxplot}
\end{figure}

First, we visualize the distribution of ratings provided by human and \ac{LLM} judges for comparative analysis, as shown in Figure~\ref{fig:human_alignment_boxplot}.
We observe that the human judges tend to provide lower and diverse ratings than the \ac{LLM} judges, indicating a potential bias in the \ac{LLM} evaluation process.

\begin{figure}
    \centering
    \includegraphics[trim=0cm 0cm 0cm 0cm, clip, width=\columnwidth]{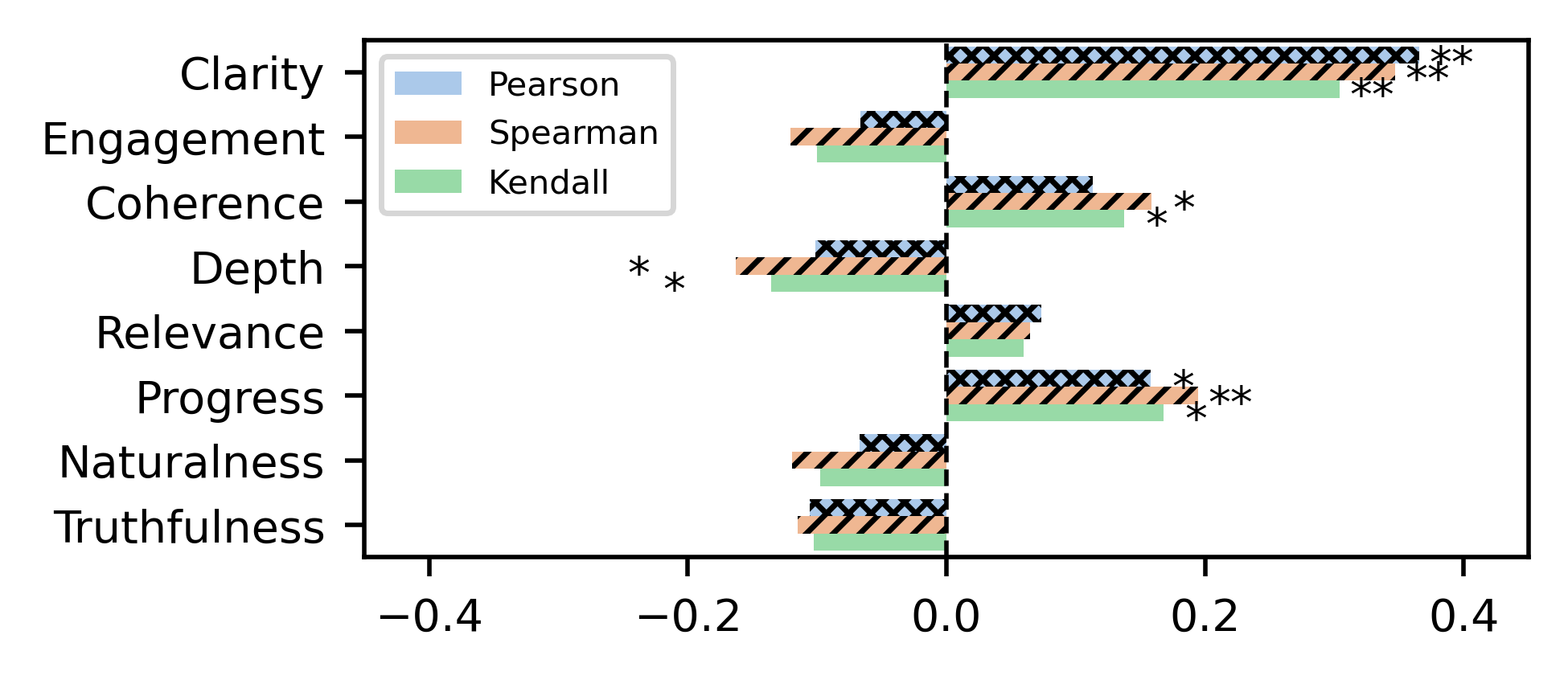}
    \caption{Human alignment correlation coefficient Pearson's $r$, Spearman's $\rho$, Kendall's $\tau$, over diverse metrics.
    Symbols denote significance test: * p<0.05, ** p<0.01.
    }
    \label{fig:Human_alignment_correlation}
\end{figure}

\begin{figure*}[t!]
    \centering
    \includegraphics[width=0.8\linewidth]{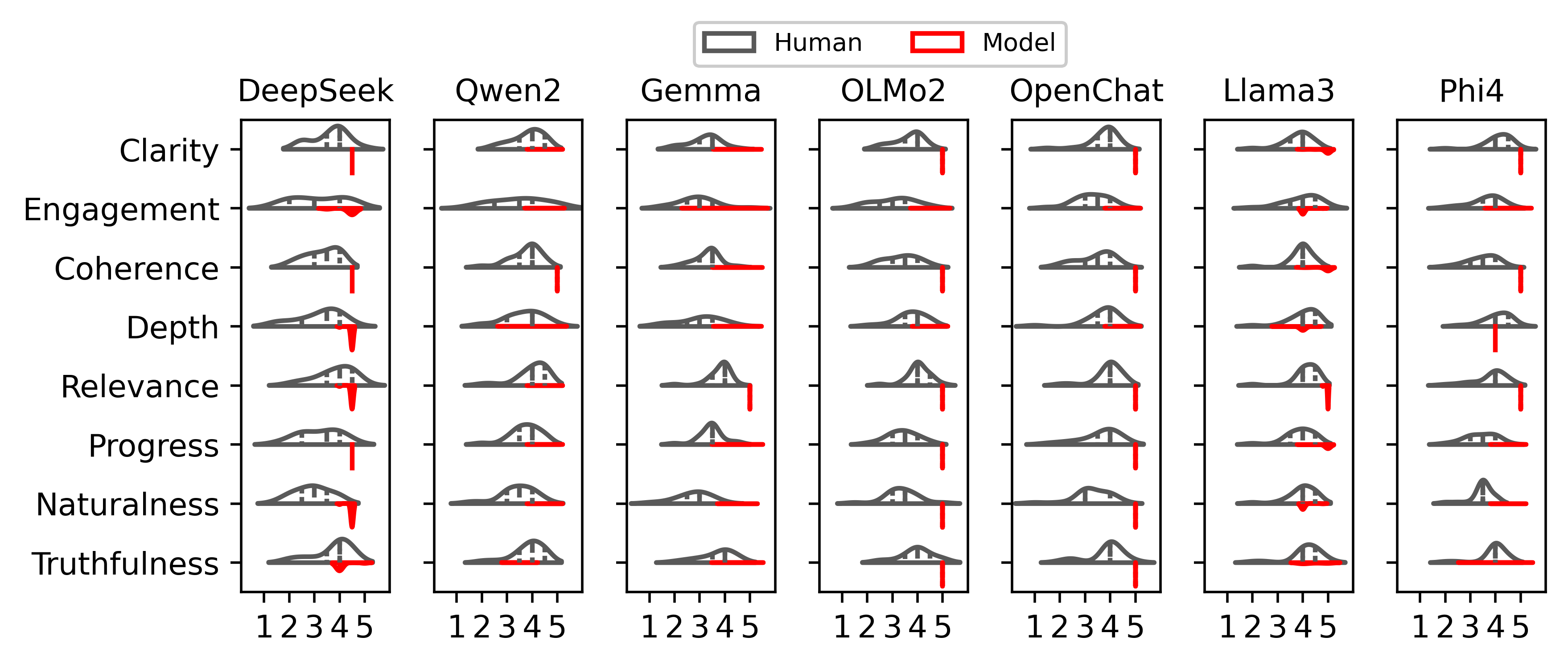}
    \caption{Distribution of model evaluations compared to human annotations across all metrics.}
    \label{fig:human_alignment_distribution}
\end{figure*}
Second, we compute the correlation coefficients between human and \ac{LLM} judges across all metrics, as shown in Figure~\ref{fig:Human_alignment_correlation}.
Metrics such as \textbf{Clarity}, \textbf{Progress}, \textbf{Coherence}, and \textbf{Relevance} exhibit higher correlations, indicating stronger alignment between \ac{LLM} and human judgments in these areas. 
This suggests that these metrics effectively capture aspects of conversational quality that resonate with human evaluators.
In contrast, metrics like \textbf{Depth}, \textbf{Engagement}, \textbf{Naturalness}, and \textbf{Truthfulness} show lower correlations, highlighting potential gaps in how \ac{LLM} judges assess these dimensions compared to human evaluators.
This misalignment may stem from the inherent difficulty of evaluating nuanced aspects such as conversational depth and authenticity, suggesting the need for improved evaluation strategies or additional calibration of \ac{LLM}-based metrics.

Third, we visualize the distribution of model evaluations compared to human annotations across all metrics, as shown in Figure~\ref{fig:human_alignment_distribution}.
We observe that LLM evaluations are tightly clustered around a high median with low variance, whereas human scores display a wider spread and a lower median. 
This suggests that LLMs tend to rate fluency and related qualities higher, possibly due to an over-reliance on surface-level text coherence.
In contrast, for metrics such as \textbf{Engagement} and \textbf{Depth}, both most LLMs and human ratings exhibit greater variability, with most LLM ratings often distributed more uniformly across the scale.
This discrepancy likely arises from fundamental differences in evaluation criteria: human judges may focus more on the quality and depth of the conversation, while LLM judges prioritize fluency and coherence.
This misalignment highlights the need for improved evaluation strategies or additional calibration of LLM-based metrics to better reflect human judgment.

\section{Discussion}

\paragraph{Implication.}
This work introduces a modular, multi-agent workflow for procedural learning that contributes to the development of scalable, LLM-based educational systems. 
The proposed large-scale dataset, grounded in structured tutorials and spanning domains, offers a valuable resource for both NLP and AI4Education communities. 
Combined with a multi-perspective evaluation protocol (automatic metrics, rubric scoring, and human assessment), the workflow enables comprehensive assessment of LLMs in educational contexts.

From the perspectives of educational AI systems,
This approach supports the development of scalable, automated tutoring systems adaptable to various topics, instructional goals, and learner profiles. By simulating interactive teacher-learner dialogues, it moves beyond static benchmarks and enables dynamic, pedagogically grounded evaluation of LLMs. This opens new directions for AI-powered curriculum design, feedback systems, and intelligent tutoring.
From a broader societal perspective, such systems can improve access to high-quality instruction, particularly in resource-limited settings. Meanwhile, their growing role also highlights the need for responsible design, transparent evaluation, and human oversight to ensure they complement rather than replace effective teaching.

\paragraph{Positioning Future Work.}
Looking ahead, several directions can further advance this work. First, incorporating real human learners and evaluators is essential to capture authentic learning dynamics and validate automated evaluation methods. Second, future research can focus on explicitly modeling pedagogical skills within LLMs, such as scaffolding, adaptive explanation, and constructive feedback, to more accurately reflect teaching competence and align LLMs with pedagogical skills. Lastly, enhancing personalization through learner modeling and adaptive dialogue strategies will improve responsiveness and realism, bringing the system closer to real-world deployment.

\section{Related Work}

\subsection{Single LLM-Agent for Education} 

LLMs have been employed as single agents in education, serving as teachers, learners, or evaluators to enhance learning. 
Teacher agents deliver knowledge, answer questions, and guide learners through structured content. 
For instance, ChatTutor~\cite{chen2024empowering} uses LLMs for course planning, adaptive quizzes, and tailored instruction via interaction, reflection, and reaction. 
Similarly, \citet{chen2024learning} show that ChatGPT as a teachable agent improves programming education by enhancing students' understanding and coding skills.
For instance, \citet{ma2024students} fine-tuned LLMs to simulate diverse language learners, aiding pre-service teachers in adapting to varied student needs.
Evaluator agents assess educational interactions and provide feedback. 
\citet{hu2025exploring} use LLMs to simulate teacher-student interactions, generate teaching reflections, and refine lesson plans, showcasing their potential to enhance pedagogy.
These roles demonstrate the versatility of LLMs in addressing educational needs, forming a foundation for multi-agent workflows.

\subsection{Multi-LLM Agent Workflows} 
Recent advancements in multi-\acp{LLM} workflows leverage agent collaboration to enhance scalability and adaptability in educational systems~\cite{chu2025llm}. 
For example, EduAgent~\cite{xu2024eduagent} integrates cognitive priors to guide reasoning in simulated interactions, while \citet{park2024empowering} align diagnostic student modeling with LLM-based tutoring through prompt engineering. 
GenMentor~\cite{wang2025mllm} dynamically tailors content to learner needs by optimizing learning paths, and LLMAgent-CK~\cite{yang2025content} employs structured multi-agent roles for content knowledge identification. 
\citet{xu2025classroom} introduce a framework with reflective and novice agents to simulate realistic student behaviors, enhancing fidelity and learning outcome assessments. 
Additionally, EvaAI~\cite{lagakis2024evaai} introduces a multi-agent framework leveraging LLMs for enhanced automated grading, further demonstrating the utility of LLMs in educational evaluation.
\citet{pei-2024-arda} propose a multi-agent framework with visual and language agents to assist multimodal LEGO assembly learning.
These approaches highlight the potential of multi-agent LLM in advancing personalized and effective educational workflows.

Compared to existing works, our approach emphasizes the integration of multiple LLMs to simulate procedural learning interactions. 
This enables a more comprehensive evaluation of pedagogical quality, instructional effectiveness, and conversational dynamics, advancing the development of scalable and adaptive educational systems.

\subsection{Educational Applications and Domains} 
Extensive work has been conducted across diverse contexts, including language learning, STEM education, and professional development. 
In language learning, studies have demonstrated the effectiveness of LLMs in enhancing student engagement through conversational interfaces~\cite{ma2024students,lagakis2024evaai}. 
In STEM education, LLMs have been applied to facilitate learning in mathematics~\citep{goutora}, physics~\citep{wang2023newton,ma2024llm}, chemistry~\citep{m2024augmenting,sprueill2024chemreasoner}, biology~\citep {huang2024protchat,ghafarollahi2024protagents}, and general scientific discovery~\citep{ma2024sciagent,baek2024researchagent}. 
Additionally, LLMs have shown promise in professional development, including applications in medical training~\citep{yang2024talk2care,abd2023large}, computer science education~\citep{zhang2024codeagent,yang2024swe,islam2024mapcoder}, and legal studies~\cite{guha2023legalbench}. 
Compared to these domain-specific applications, our work leverages large-scale learning materials spanning a wide range of domains, enabling a more comprehensive exploration of procedural learning and pedagogic quality assessment.

\section{Conclusion}
In this work, we introduced a multi-LLM agent workflow for simulating procedural learning and assessing pedagogic quality at scale. 
Our framework leverages teacher, learner, and evaluator agents to enable dynamic teaching-learning simulations grounded in diverse instructional content. 
We presented a large-scale dataset and a comprehensive evaluation protocol, demonstrating the workflow's effectiveness across various setups. 
We also identify areas for improvement, such as enhancing learner engagement and aligning automated evaluations with human judgments. 
This work lays the foundation for scalable, adaptive educational systems, bridging NLP with AI4Education.


\section*{Limitations}

While our study presents a large-scale, structured framework for simulating procedural teaching-learning interactions, there remain two areas with potential for further enhancement.
First, the learner agents are simulated using LLMs within a predefined interaction flow. This setup enables controlled and consistent experimentation but may not fully reflect the variability and learning behavior of real human learners. Future work could incorporate adaptive dialogue policies or human-in-the-loop learners to improve the realism and responsiveness of interactions.
Second, although our dataset spans 17 domains and 727 topics, variation in topic complexity and data distribution could influence performance comparisons. Certain domains may be overrepresented or inherently more challenging, which could affect model generalizability. Future extensions can include domain-balanced evaluation and fine-grained analysis at the topic level to better understand model behavior across different knowledge structures.
These considerations do not compromise the core contributions of our work but highlight opportunities for expanding its applicability and pedagogical fidelity.

\bibliography{custom}

\appendix

\clearpage
\section{Implementation Details}
The implementation of the workflow is based on LangGraph.~\footnote{\url{https://langchain-ai.github.io/langgraph/}}
We utilize 7 the most capable openly available \acp{LLM} hosted on Ollama~\footnote{\url{https://ollama.com/search}} as the backbone of the proposed workflow:
\begin{enumerate*}[leftmargin=*, label=(\arabic*)]
    \item DeepSeek: A 7B parameter, bilingual-supportive, open-source model from DeepSeek.
    \item Qwen2: A 7B parameter, multilingual-supportive, open-source model from Alibaba.
    \item Gemma: A 7B parameter, lightweight, open-source model developed by Google DeepMind.
    \item OLMo2: A 7B parameter, open-source model from AllenAI, demonstrating competitive performance with equivalently sized open models.
    \item OpenChat: A 7B parameter, open-source model from Tsinghua, surpassing ChatGPT 3.5 on various benchmarks.
    \item Llama3: An 8B parameter, highly capable openly available model from Meta.
    \item Phi4: A 14B parameter, open-source model from Microsoft.
\end{enumerate*}
It is possible to seamlessly switch to other backbone models as needed.
We use GPT-4, the closed-source model with 1.76T parameters from OpenAI, by calling the API of the system~\cite{hurst2024gpt}.
We choose the sentence transformer model all-MiniLM-L6-v2~\cite{reimers2019} as the embedding model for semantic similarity search. 

We set the maximum number of tokens for each model to 4096, and the maximum number of interactions to 40.
We set the temperature to 0.0 to ensure deterministic outputs.
The default batch size for model inference is set to 16. 
All experiments are conducted on NVIDIA H100 (94GB) and NVIDIA A100 SXM4 (40GB) GPUs. 
Additionally, our workflow is compatible with deployment on CPUs. 

\section{Prompt Templates}\label{sec:prompt}
\begin{table}[h!]
\centering
\resizebox{\columnwidth}{!}{
\setlength{\tabcolsep}{1pt} 
\begin{tabular}{@{}cl@{}}
\toprule
\textbf{Agent} & \textbf{Prompt Template}\\ \midrule
Teacher        & 
\begin{tabular}[c]{@{}l@{}}
You are an expert teacher guiding a learner step by step through\\ a tutorial via multi-turn conversations.\\ 
\textbf{Your role}: \\
(1) Answer learner questions related to the instructions as the\\\ first priority. \\ 
(2) If the learner requests to move on without asking questions,\\ provide the current step’s instruction directly. \\
(3) Once the final step has been completed and the learner has\\ no further questions, acknowledge completion by appending the\\ token `FINISHED' to your response. \\
(4) Highlight key phrases in your response if they appear in the\\ tutorial.\\ 
Respond in character as a teacher — no system-level messages\\ or meta-comments. 
Keep responses short and focused.\\ 
\textbf{Tutorial Summary}: \{summary\}\\
\textbf{Step} \{current\_step\_index\} \{current\_step\_content\}\\ 
\textbf{Learner}: \{user\_utterance\}; \textbf{Teacher}:\end{tabular} \\ \midrule
Learner        & 
\begin{tabular}[c]{@{}l@{}}
You are a curious student engaged in a step-by-step learning\\ conversation with a teacher.\\ 
\textbf{Your role}: \\
(1) Carefully read and understand the teacher’s instructions. \\
(2) Respond naturally, politely, and concisely.\\ 
(3) If the instruction is unclear, ask a brief and specific\\ question to clarify it.\\ 
(4) If the instruction is clear, acknowledge it briefly and\\ politely ask to move on to the next step.\\ 
(5) If the teacher opens with a BEGIN message, just acknowledge\\ it and do not ask questions.\\ 
(6) If the teacher says `FINISHED' or indicates that the tutorial\\ is over, thank them politely and say nothing more.         \\ 
Respond in character as a student — no system-level messages\\ or meta-comments. Keep responses short and focused.\\ 
\textbf{Teacher}: \{instruction\}; \textbf{Learner}:\end{tabular}                     \\ \midrule
Evaluator      & \begin{tabular}[c]{@{}l@{}}
Evaluate the following teacher-learner conversation: 
\{conversation\} \\
{[}\textbf{Reference}: \{tutorial\}{]}; \\
\textbf{Score} from 1 to 5: \{criterion\_name: criterion\_rubric\} \\
Provide scores in this format: \{criterion\_name: rubric\_score\}\end{tabular}  \\ \bottomrule
\end{tabular}}
\caption{Prompt templates used in this work.}\label{tab:prompt_template}
\end{table}
As shown in Table~\ref{tab:prompt_template}, we provide the prompt templates for each step of the workflow.
The prompt templates are designed to be flexible and can be easily adapted to different tasks or domains.
The templates include placeholders for input data, which can be replaced with specific examples or parameters as needed.
The templates also include instructions for the model, guiding it on how to process the input data and generate the desired output.
The templates are structured to ensure that the model understands the context and requirements of each task, facilitating effective communication between the user and the model.

\section{Number of Articles across Domains}\label{sec:article_per_domain}
\begin{table*}[h!]
\centering
\scriptsize
\resizebox{\textwidth}{!}{
\begin{tabular}{@{}clclclclclc@{}}
\toprule
\textbf{ID} & \textbf{Domain}         & \textbf{\#Article} & \textbf{} & \textbf{ID} & \textbf{Domain}         & \textbf{\#Article} & \textbf{} & \textbf{ID} & \textbf{Domain}   & \textbf{\#Article} \\ 
\midrule
1            & Automobiles \& Vehicles & 166                &           & 
7            & Family \& Parenting     & 150                &           & 
13           & Pets \& Animals   & 131                \\
2            & Business \& Finance     & 98                 &           & 
8            & Food \& Cooking         & 127                &           & 
14           & Relationships     & 368                \\
3            & Career \& Workplace     & 139                &           & 
9            & Health                  & 363                &           & 
15           & Self-Improvement  & 222                \\
4            & Culture \& Lifestyle    & 197                &           & 
10           & Hobbies \& Leisure      & 306                &           & 
16           & Technology        & 227                \\
5            & Education               & 296                &           & 
11           & Home \& Garden          & 266                &           & 
17           & Travel \& Leisure & 61                 \\
6            & Entertainment           & 209                &           & 
12           & Personal Care \& Beauty & 358                &           &              &                   &                    \\ \bottomrule
\end{tabular}
}
\caption{Number of WikiHow articles across 17 domains.}\label{tab:article_per_domain}
\end{table*}
As shown in Table~\ref{tab:article_per_domain}, we provide the number of articles across different domains.
The table includes the total number of articles, the number of unique topics, and the average number of articles per topic for each domain.
The data is collected from the WikiHow dataset, which contains a diverse range of articles covering various topics.
The domains include health, technology, lifestyle, and more.
The number of articles in each domain varies significantly, with some domains having a larger number of articles than others.
The table provides a comprehensive overview of the distribution of articles across different domains, highlighting the diversity and breadth of topics covered in the WikiHow dataset.

\section{Detailed Rubrics for Evaluation}\label{sec:appendix:rubric}
\begin{table*}[htb!]
\centering
\tiny
\resizebox{\textwidth}{!}{
\begin{tabular}{@{}ll@{}}
\toprule
\textbf{Criterion} & \textbf{Rubric} (5 -- Excellent, 4 -- Good, 3 --Average, 2 -- Poor, 1 -- Very Poor) \\ \midrule                                                                  
Clarity$^{\dagger}$            & \begin{tabular}[c]{@{}l@{}}Is the teacher’s instruction clear, well-structured, and easy to understand?\\ 5 -- The instruction is highly clear, logically structured, and easy to understand. No ambiguity, and key concepts are well explained.     \\ 
4 -- The instruction is mostly clear, with minor ambiguities or areas that could be refined. The overall structure is logical.          \\ 
3 -- The instruction is somewhat clear, but there are noticeable ambiguities or structural issues that may cause some confusion.     \\ 
2 -- The instruction is difficult to follow due to unclear phrasing, missing details, or a lack of logical flow. Learners may struggle to understand.\\ 
1 -- The instruction is highly unclear, disorganized, or misleading. Learners are likely to be confused or misunderstand key points.\end{tabular}                                \\ \midrule
Truthfulness$^{\dagger}$       & \begin{tabular}[c]{@{}l@{}}Does the generated response stay within the scope of the reference tutorial?\\ 
5 -- All information is completely accurate, directly aligned with the given tutorial or trusted sources. No hallucinations or misleading details.          \\ 
4 -- The response is largely correct but may contain small factual errors, slightly ambiguous statements, or minor misinterpretations.     \\ 
3 -- Includes some correct information but also contains notable factual errors, misrepresentations, or unsupported claims.        \\ 
2 -- A majority of the response is incorrect or misleading, with only a small portion being factually accurate.  \\ 
1 -- The response is entirely false, misleading, or contradicts known facts or the provided tutorial.\end{tabular} 
\\ \midrule
Engagement$^{\ddagger}$         & \begin{tabular}[c]{@{}l@{}}Does the learner actively participate by asking meaningful questions or responding thoughtfully?\\ 
5 -- The learner frequently asks insightful questions, provides thoughtful responses, and demonstrates strong engagement.\\ 
4 -- The learner engages well, asking relevant questions and responding appropriately, with only minor missed opportunities.     \\ 
3 -- The learner participates but does not consistently ask meaningful questions or provide detailed responses.       \\ 
2 -- The learner shows minimal engagement, rarely asking questions or contributing meaningfully to the discussion.        \\ 
1 -- The learner is disengaged, does not ask questions, and provides minimal or no meaningful responses.\end{tabular}                                                              \\ \midrule
Coherence$^{\star}$          & \begin{tabular}[c]{@{}l@{}}Is the conversation logically structured, with smooth transitions between steps?\\ 
5 -- The conversation is well-structured, with smooth transitions and a logical flow from one step to the next.\\ 
4 -- The conversation is mostly coherent, with only minor inconsistencies or abrupt transitions.       \\ 
3 -- The conversation has some structural issues, with occasional jumps or awkward transitions.       \\ 
2 -- The conversation lacks logical flow, with multiple abrupt or confusing transitions. \\ 
1 -- The conversation is highly disorganized, making it difficult to follow the sequence of steps.\end{tabular}                              \\ \midrule
Depth$^{\star}$              & \begin{tabular}[c]{@{}l@{}}Does the conversation go beyond surface-level discussion and explore concepts in sufficient detail?\\ 
5 -- The conversation explores concepts in-depth, with thorough explanations and critical thinking.\\ 
4 -- The conversation provides mostly in-depth discussion, but some areas could be explored further.\\ 
3 -- The conversation touches on key concepts but lacks depth in explanations or follow-up discussions.\\ 
2 -- The conversation is mostly superficial, with little exploration beyond basic information.\\ 
1 -- The conversation is highly shallow, lacking any meaningful discussion or explanation.\end{tabular}                                   \\ \midrule
Relevance$^{\star}$          & \begin{tabular}[c]{@{}l@{}}Do the responses stay on-topic and align with the tutorial’s instructions and context?\\ 
5 -- The responses are consistently relevant and closely follow the tutorial's instructions and context.\\ 
4 -- The responses are mostly relevant, with only occasional minor deviations.\\ 
3 -- The responses are somewhat relevant but occasionally stray off-topic.\\ 
2 -- The conversation frequently goes off-topic, with limited alignment to the tutorial's instructions.  \\ 
1 -- The responses are largely irrelevant, making it difficult to follow the intended learning path.\end{tabular}                                                                   \\ \midrule
Progress$^{\star}$           & \begin{tabular}[c]{@{}l@{}}Does the conversation effectively move forward through the tutorial steps in a logical manner?\\ 
5 -- The conversation progresses efficiently, covering each step logically and without unnecessary repetition.\\ 
4 -- The conversation generally moves forward well, with only minor delays or repetitions.\\ 
3 -- The conversation progresses but sometimes gets stuck or moves inefficiently.\\ 
2 -- The conversation struggles to move forward, often repeating steps or getting sidetracked.\\ 
1 -- The conversation lacks clear progress, frequently revisiting previous steps or failing to complete the tutorial.\end{tabular}                                    \\ \midrule
Naturalness$^{\star}$        & \begin{tabular}[c]{@{}l@{}}Does the conversation feel fluid and human-like, avoiding robotic or overly scripted responses?\\ 
5 -- The conversation feels natural and human-like, with engaging and varied responses.\\ 
4 -- The conversation is mostly natural, but some responses feel slightly mechanical or repetitive.\\ 
3 -- The conversation has a mix of natural and robotic responses, with some unnatural phrasing.        \\ 
2 -- The conversation often feels artificial or scripted, with little variation in responses.            \\ 
1 -- The conversation is highly robotic or formulaic, making it feel unnatural and disengaging.\end{tabular}                                      \\ \bottomrule
\end{tabular}
}
\caption{Rubric for \acp{LLM} and human annotators.  Symbols $^{\dagger}$, $^{\ddagger}$, and $^{\star}$ represent the target evaluation role as teacher, learner, and conversation, respectively.}\label{tab:rubric}
\end{table*}
As shown in Table~\ref{tab:rubric}, we provide the detailed rubrics for evaluation.
The rubrics are designed to assess the quality and effectiveness of the generated articles, used for both LLM and human judges.
Each criterion is defined with specific guidelines to ensure consistent and objective evaluation.
To be more specific, we provide three types of rubrics tailored to the target evaluation roles: teacher, learner, and conversation.
For the teacher role, the rubric emphasizes Clarity and Truthfulness, ensuring that the generated teaching material is accurate, tutorial-based, and easy to comprehend.
For the learner role, the rubric emphasizes Engagement, encouraging learners to actively participate by asking meaningful questions and providing thoughtful responses.
For the conversation role, the remaining rubrics assess the quality of the conversation in terms of the learning process (Progress), logical flow (Relevance and Coherence), and linguistic features (Naturalness).

\section{Knowledge Graph for WikiHow Articles}\label{sec:appendix:kg}
\begin{figure}[t!]
    \centering
    \includegraphics[width=1\linewidth]{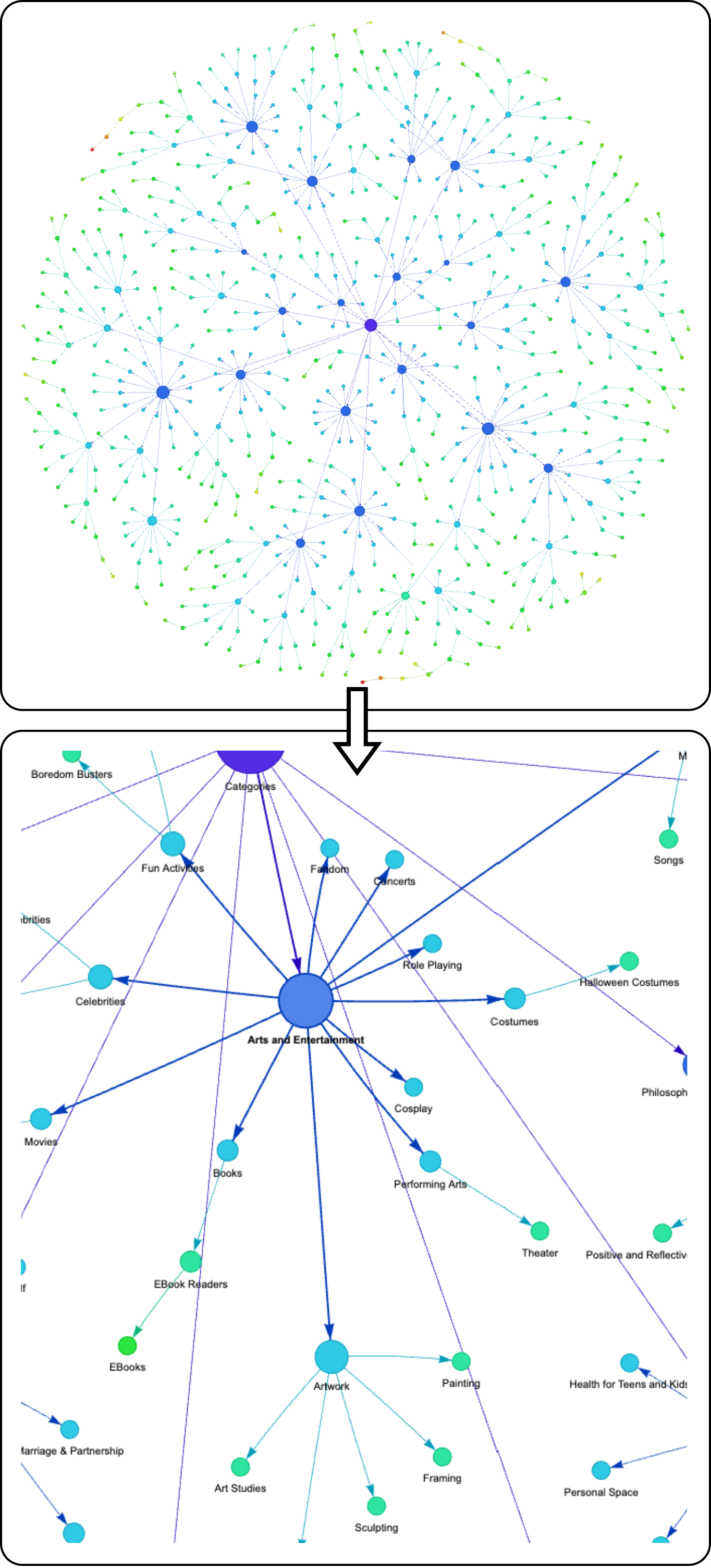}
    \caption{Visualization of the knowledge graph composed from WikiHow articles with hierarchical categories.}
    \label{fig:enter-label}
\end{figure}

As shown in Figure~\ref{fig:enter-label}, we provide the knowledge graph for WikiHow articles.
The graph illustrates the hierarchical relationships among domains, topics, and articles within the dataset, offering a comprehensive visual representation of the knowledge structure. 
The root node serves as a dummy root category, with first-level nodes representing domains, intermediate nodes denoting hierarchical topics, and leaf nodes corresponding to individual articles.
The edges represent the relationships between them.

The knowledge graph serves as a valuable resource for understanding the structure and organization of information within the WikiHow dataset.
It brings a broader implication of the proposed dataset, which can be used for various applications, including information retrieval, recommendation systems, and knowledge discovery.

\section{Case Study}
\begin{table*}[htb!]
\centering
\tiny
\resizebox{\textwidth}{!}{
\setlength{\tabcolsep}{1pt} %
\begin{tabular}{@{}l@{}}
\toprule
\textbf{Tutorial: Part 2 of 3:Writing the Body of the Speech 
}\\ 
\begin{tabular}[c]{@{}l@{}}
\textbf{Categories:} Education and Communications \textgreater{}\textgreater Personal Development \textgreater{}\textgreater Maturity \textgreater{}\textgreater Leadership \textgreater{}\textgreater School Leadership\end{tabular} \\
\midrule
\begin{tabular}[c]{@{}l@{}}
\underline{\textbf{Step 1:} State your main ideas on how to improve the school.} \\
You should have at least three ideas you feel would benefit your school and classmates. 
This gives your classmates the incentive to vote for you and shows you want the position as an\\ opportunity to help others.\\ 
(1) You should \colorbox{lightpurple}{list your ideas and then expand on them} later in the body. It might take a bit of research to figure out what you want to change. 
Ask around the school, talking to students\\ and teachers, and see where there's room for improvement. What are the concerns of the students?
What are people happy with regarding the school? What would they like to see change? \\
Asking these questions can help you get a sense of your audience and community.\\ 
(2) Remember, you should not make promises you cannot keep. Do not say anything just to get elected.
While many students might want gum-chewing policies eliminated or for the lunch\\ period to run twice as long, this is probably not necessary or possible.
Try to focus on areas that seem important to keep your school running safely and efficiently. 
Concerns about things\\ like bullying, academic standards, and extracurricular activities should be your concern  
 over fun and games. \\ 
(3) A good opening statement for your body would \colorbox{pink}{state the causes important to you and what you plan to do about them}. 
For example, if you were running for president, you could say\\ something like, "I understand we need to improve how we handle \colorbox{green}{bullying}, increase interest in \colorbox{blue}{extracurricular activities}, and expand \colorbox{orange}{access to AP courses} throughout the school. \\
As your president, I would work to bring in speakers to talk about sensitivity in the classroom, increase advertising for basketball games and quiz bowl tournaments, and start a tutoring\\ program to help students struggling with certain subjects."\end{tabular} 
\\
\begin{tabular}[c]{@{}l@{}}
\underline{\textbf{Step 2:} Find support for those ideas.}\\ 
You should do a little \colorbox{yellow}{extra research} in addition to talking to classmates and teachers. \colorbox{lightgreen}{Have some specific plans on how you would enact change} in your school. \\(1) Using the school library or computers, figure out the best means to tackle certain problems many schools face.  
\colorbox{yellow}{How have other} \colorbox{yellow}{schools dealt with} bullying? Poor test scores?\\ 
Low interest in extracurricular activities? What can you reasonably do as a student council member to address these problems? \\
(2) You do not have to have a point-by-point plan laid out, but a few sentences on some preliminary ideas can help you stand out from your peers. 
People are more inclined to vote for\\ someone who's thought about how to solve problems in addition to identifying problems.\end{tabular}
\\
\begin{tabular}[c]{@{}l@{}}
\underline{\textbf{Step 3:} Keep your ideas short but very strongly worded.}\\ 
\colorbox{lightblue}{Your body should be about two paragraphs of 5 to 6 sentences each.} This might sound brief, considering how much information you'll have to get across, but you have limited time and\\ must keep people's attention.
It can help to write more than you need and gradually cut down your speech to the bare essentials.
This will help it to be shorter meaning that students won't\\ get bored when listening to your speech.\end{tabular}\\
\midrule
\begin{tabular}[c]{@{}l@{}}\textbf{Generated Conversation:}\\ 
\underline{\textbf{Teacher:}} Let's begin: \underline{Part 2 of 3:Writing the Body of the Speech}. \textsc{begin}\\ \underline{\textbf{Learner:}} Okay, I'm ready to start learning about writing the body of the speech. Please guide me through this part.\\ \underline{\textbf{Teacher:}} Great! For the body of your speech, you'll want to \colorbox{lightpurple}{expand on the main ideas you've listed in \textbf{step 1}}. \\ 
Each idea should have its own paragraph. For example:\\ 
\colorbox{green}{\textit{Bullying}}: 
\colorbox{pink}{
Explain why addressing bullying is crucial and how it impacts students' well-being. Discuss specific strategies} like bringing in sensitivity speakers or increasing awareness\\ through workshops.\\ 
\colorbox{blue}{\textit{Extracurricular Activities}}:
\colorbox{pink}{Highlight the importance of a diverse range of activities for student engagement and personal growth. 
Outline plans} to 
boost participation, such as improving\\ advertising methods or creating more opportunities for clubs and sports teams.
\\
\colorbox{orange}{\textit{AP Courses Access}}: 
\colorbox{pink}{Address how access to Advanced Placement courses can benefit students' academic futures.
Propose solutions} like forming partnerships with local universities for\\ guest lectures or establishing a tutoring program.
Each paragraph should start with a clear topic sentence that directly relates to one of your main ideas, followed by supporting sentences\\ that detail your plans and why they're beneficial. 
This will make your speech compelling and informative.\\ 
\underline{\textbf{Learner:}} Thank you, teacher. I understand the instructions. Let's begin writing the body of my speech.\\ 
\colorbox{green}{\textit{Bullying}}:  Addressing bullying is \colorbox{pink}{crucial because it significantly impacts} students' mental health and academic performance. 
By implementing sensitivity speakers who share personal\\ experiences or insights on empathy and understanding, we can foster a more inclusive environment where students feel safe and supported. 
Additionally, organizing workshops that educate\\ both students and staff about the effects of bullying will help in creating awareness and promoting a culture of respect.\\ 
\colorbox{blue}{\textit{Extracurricular Activities}}: A \colorbox{pink}{diverse range of activities is essential} for student engagement and personal growth. To boost participation, we can improve advertising methods by\\ utilizing social media platforms to reach out to students more effectively. 
We should also create more opportunities for clubs and sports teams by allocating additional funds specifically\\ for these initiatives or seeking sponsorships from local businesses. 
This will not only increase student involvement but also enhance their leadership skills.\\ 
\colorbox{orange}{\textit{AP Courses Access}}:  \colorbox{pink}{Access to Advanced Placement courses is vital as it prepares students for college-level work and future academic pursuits.} 
By forming partnerships with local\\ universities, we can invite guest lecturers who offer insights into advanced topics or provide guidance on college applications. 
Establishing a tutoring program will ensure that all students\\ have the opportunity to excel in AP courses by receiving personalized assistance from peers or teachers. This approach not only enriches their educational experience but also boosts their\\ confidence and academic readiness for higher education. 
I believe these paragraphs effectively address each main idea with clear topic sentences, supporting details, and explanations of\\ why these strategies are beneficial.         \\ \underline{\textbf{Teacher:}} Great job! Your speech outline covers the key issues comprehensively. 
Make sure to \colorbox{lightgreen}{include a personal touch by sharing your ideas on how you would implement these changes} \\
if elected. This will make your speech more relatable and engaging for your peers.  
Next, consider \colorbox{yellow}{adding some statistics or examples from other schools}that have successfully addressed\\  these problems. 
This will strengthen your argument and show that you've done \colorbox{yellow}{thorough research}. Once you're ready to move on, let me know! \\ \underline{\textbf{Learner:}} Thank you for the feedback! \colorbox{lightgreen}{I'll incorporate my personal experiences and find relevant statistics.} When I'm finished with this step, please let me know.\\ \underline{\textbf{Teacher:}} Great job focusing on personal experiences and stats! For \textbf{Step 3}, concentrate on \colorbox{lightblue}{crafting a concise body that packs a punch. Aim for two paragraphs of about 5 to 6 sentences each.} \\
Keep your ideas strong and impactful.  Remember, you have limited time, so it's crucial to be succinct and engaging. 
Write more than necessary initially, then refine your speech to its essentials. \\ 
This will ensure your message is clear and keeps the audience's attention. Once you've completed this step, let me know!\\ \underline{\textbf{Learner:}} Understood, Teacher. I'll focus on \colorbox{lightblue}{creating a concise body with impactful ideas} for my speech. 
\colorbox{lightblue}{I plan to write two paragraphs of about 5 to 6 sentences each,}ensuring they are \\ strong and engaging.  
After drafting, I will refine the text to its essentials to maintain clarity and audience interest. 
Once done, I'll notify you. Thank you!\\ 
\underline{\textbf{Teacher:}} We've reached the end of the tutorial. Bye! \textsc{finished}\\ 
\underline{\textbf{Learner:}} Thank you for the tutorial, I've learned a lot. Goodbye!\end{tabular} \\ \bottomrule
\end{tabular}
}
\caption{An example of a generated conversation with the reference tutorial. Text highlighted in the same color indicates relevant parts, as marked by human annotation, and is intended to improve human readability.}\label{tab:case_study}
\end{table*}
As shown in Table~\ref{tab:case_study}, we provide a case study of the generated conversations.
The case study includes an example of educational conversation generated by the proposed workflow, grounded on a WikiHow tutorial spanning a hierarchical range of topics. 
Highlighted texts in the same color indicate relevant parts, as marked by human annotation, to enhance readability and facilitate a better understanding of the generated content.
This demonstrates a nuanced understanding of diverse topics and the ability to generate coherent, contextually relevant content.

\section{Risks, Ethics, Data and AI Tool Usage}

We recognize that the use of \acp{LLM} in educational contexts presents several risks, including the potential for misinformation, bias, and over-reliance on automated tools. 

The proposed workflow is intended to assist users in generating educational content and supporting learning—not to replace human educators or learners. 
Rather, it aims to augment human capabilities and provide supplementary resources to enhance the educational experience.

The WikiHow articles used in this work are publicly available and licensed under the Creative Commons Attribution-NonCommercial-ShareAlike 3.0 Unported License~\url{https://creativecommons.org/licenses/by-nc-sa/3.0/}. 

This dataset is used solely for research purposes, with no intention of commercial use.
It does not contain personally identifying information or offensive content.

We acknowledge the ethical challenges associated with deploying \acp{LLM} in education, particularly the risk of generating biased, inappropriate, or misleading content. 
To mitigate these concerns, we are committed to ensuring that the content generated through our workflow is accurate, reliable, respectful, and free from harmful bias.

In line with our ethical standards, we limited the use of AI tools (e.g., ChatGPT, Copilot, Grammarly) to surface-level tasks such as grammar checking and stylistic refinement, always based on our original writing. 
No AI tools were used to generate original content or ideas in this paper.

\end{document}